\documentclass[3p,twocolumn]{elsarticle}

\usepackage{amsmath,amssymb,amsfonts}
\usepackage{graphicx}
\usepackage{subfigure}
\usepackage{textcomp}
\usepackage{algorithmicx,algorithm}
\usepackage{algpseudocode}
\usepackage{array}
\usepackage{booktabs}
\usepackage{appendix}
\usepackage{lineno,hyperref}
\modulolinenumbers[5]
\journal{Journal of \LaTeX\ Templates}

\begin{document}

\begin{frontmatter}

\title{Conditional Image Generation with One-Vs-All Classifier}

\author[mymainaddress]{Xiangrui Xu}
\author[mymainaddres]{Xiangrui Xu}
\ead[url]{xiangruixu7@163.com}

\author[mymainaddress]{Yaqin Li}
\ead[url]{leeyaqin@whpu.edu.cn}

\author[mymainaddress]{Cao Yuan\corref{mycorrespondingauthor}}
\cortext[mycorrespondingauthor]{Corresponding author}
\ead{yc@whpu.edu.cn}

\address[mymainaddress]{School of Mathematics and Computer Science, Wuhan Polytechnic University, Hubei China}
\address[mymainaddres]{School of}

\begin{abstract}
This paper explores conditional image generation with a One-Vs-All classifier based on the Generative Adversarial Networks (GANs). Instead of the real/fake discriminator used in vanilla GANs, we propose to extend the discriminator to a One-Vs-All classifier (GAN-OVA) that can distinguish each input data to its category label. Specifically, we feed certain additional information as conditions to the generator and take the discriminator as a One-Vs-All classifier to identify each conditional category. Our model can be applied to different divergence or distances used to define the objective function, such as Jensen-Shannon divergence and Earth-Mover (or called Wasserstein-1) distance. We evaluate GAN-OVAs on MNIST and CelebA-HQ datasets, and the experimental results show that GAN-OVAs make progress toward stable training over regular conditional GANs. Furthermore, GAN-OVAs effectively accelerate the generation process of different classes and improves generation quality.
\end{abstract}


\end{frontmatter}

\section{Introduction}
\begin{figure*}[htbp]
\centering
\includegraphics[scale=0.6]{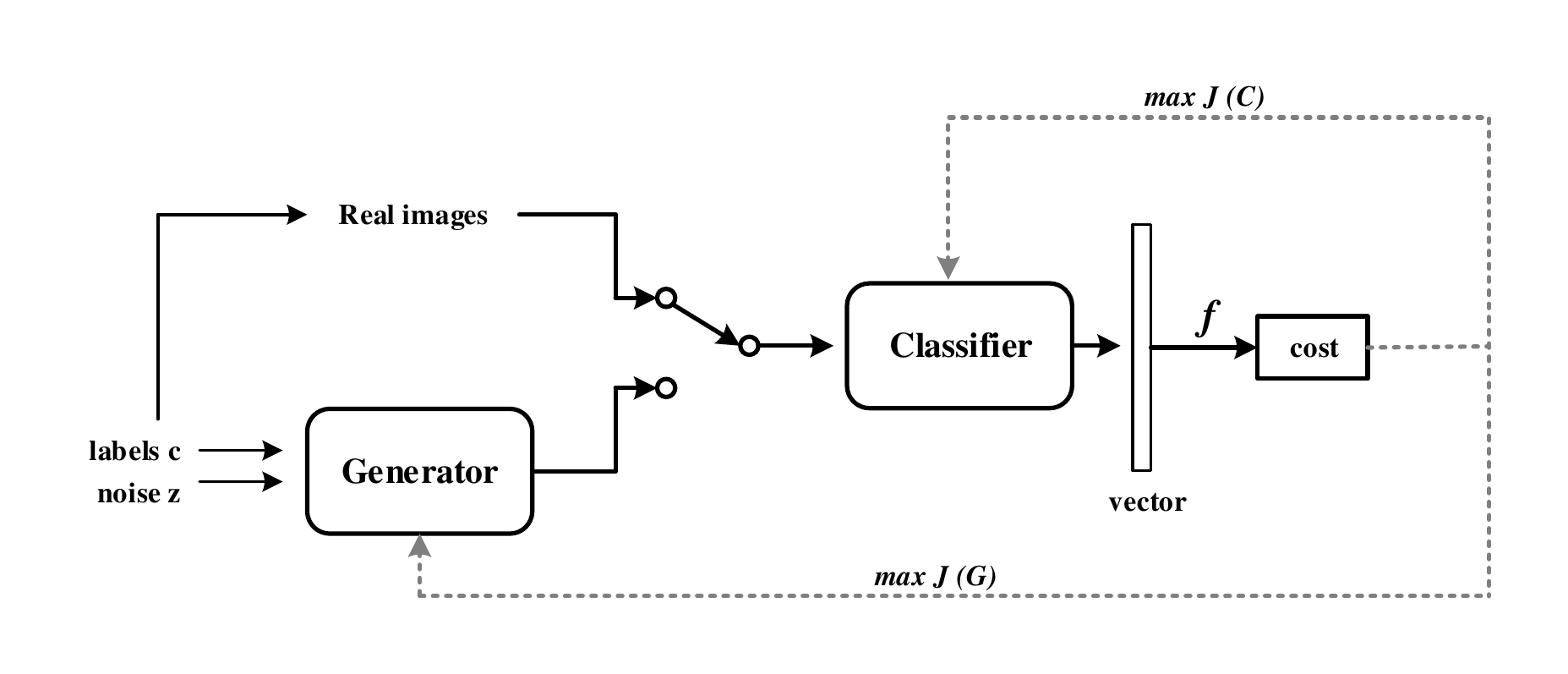}
\caption{Workflow of GAN-OVA.}
\label{fig1}
\end{figure*}

Since Ian Goodfellow proposed Generative Adversarial Networks (GAN) in 2014 \cite{goodfellow2014generative}, GANs have become a hot and attractive topic in generative models training \cite{ma2019fusiongan,wang2018graphgan,salimans2016improved}. Recent advances on GANs \cite{mirza2014conditional,odena2017conditional,van2016conditional} have devoted to building a series of conditional versions of GANs, which attempts better to guide data generation process under certain additional information. Condition-based image synthesis is important for many applications, such as image synthesis and editing \cite{zhang2017stackgan,reed2016generative,perarnau2016invertible}, speech enhancement \cite{michelsanti2017conditional}, and object detection\cite{nguyen2017shadow}, etc. \cite{li2015iris,zhang2017age,isola2017image}.

The pioneer method to control the generation process is Conditional Generative Adversarial Network (CGAN) \cite{mirza2014conditional}, where the generator and discriminator are both feeds by the class labels as condition input. Because CGAN has no conditional constraints on the generated data, the conditional information may be ignored.  Recent advances in conditional image generation focus on adding auxiliary classifier for the discriminator, such as ACGAN \cite{odena2017conditional}. The auxiliary classifier is along with a specialized cost function over the class labels, so that generator can utilize the additional condition to produce a specific category sample. However, both synthetic images and real images share the same category labels, which may confuse the auxiliary classifier.

Amongst these existing methods, the discriminator only served as a binary classifier to distinguish between real data and generated data. In the computer vision community, however, discriminative models have performed excellent results in many stream tasks \cite{ma2019image,yi2019multi,cui2018subspace,xu2020identity}, including multi-classification. Therefore, it is natural to consider whether to carry the discriminators in GANs over the multiclass situation.

Motivated by such an intuition, we propose to expand the previous real/fake classifier of the discriminator to a multiclass classifier \cite{rifkin2004defense} (One-Vs-All classifier), called GAN-OVA, to distinguish different conditional samples from real and synthesized images. We note that our GAN-OVA can be applied to different divergence driving methods with different target labels assigned. We explore the relation of GAN-OVA with Jensen-Shannon (JS) divergence \cite{goodfellow2014generative} and Earth-Mover (EM) (or called Wasserstein-1) distance \cite{martin2017wasserstein}. In JS divergence-driven GAN-OVA, in addition to the class labels of real data, a 'fake' class for generating data is added, which can effectively avoid confusion caused by using the same category label on both synthetic images and real images. The generator network in GAN-OVA is the same as other conditional GANs, which takes noises and additional conditions as input to capture the conditional data distribution. Cooperation between the One-Vs-All classifier and the conditioned generator can direct the synthesis of specific samples.

We evaluate GAN-OVA on MNIST \cite{lecun1998gradient} and CelebA-HQ \cite{karras2017progressive} datasets and use ACGAN for comparison. Compared with regular conditional GAN, GAN-OVA mainly derives two benefits. First, the One-Vs-All classifier effectively improves the model stability. Second, GAN-OVA can effectively accelerate the generation process of different classes and improve generation quality. Additionally, we explore the interpretable representation \cite{chen2016infogan} between the condition variables and the observations. Experiments show that our adversarial pair learns the feature representations from condition variables.

The structure of the rest of the paper is as follows. Section \ref{Background} provides two different methods of generative adversary nets as the background. Then we formally set out the GAN-OVA in Section \ref{GAN-OVA}. Experiments are shown in Section \ref{Experiments} with a conclusion and discussion in Section \ref{Conclusion and discussion}

\section{Background}
\label{Background}
In the GAN mechanism, the generator network defines a probability distribution $p_{g}$ while the discriminator calculates and minimizes the divergence between $p_{g}$ and the real data distribution $p_{r}$ so that realize the realistic image generation. In this section, we discuss two types of divergences used to define the objective function, that is, JS divergence and EM distance. We call the models driven by these two methods as JS divergence-driven GAN and EM distance-driven GAN, respectively.
%-------------------------------------------------------------------------
\subsection{JS divergence-driven GAN}
GANs estimate a generative model via an adversarial learning process, in which a generator network $G$ and a discriminator network $D$ are trained simultaneously.

In JS divergence-driven GANs, $G's$ goal is to build a distribution $p_{g}$ close to the real data distribution $p_{r}$ over real data $x$ by transforming a prior noise distribution $ p_{z}(z)$ into a data space $G(z;\theta_{g})$. And Discriminator$ D(x; \theta_{d})$ targets to recognize whether an image from $p_{g}$ or $p_{r}$. Formally, the value function $V(D,G)$ is described as follows:
\begin{equation}
\begin{split}
 \min_{G} \max_{D}V(D,G)= \mathbb{E}_{x\sim p_{r}(x)}[\log D(x)] +  \\
 \mathbb{E}_{z\sim p_{z}(z)}[\log (1-D(G(z)))]\label{eq1}
\end{split}
\end{equation}

The minimization value of function $V(D,G)$ is equal to the minimization JS divergence between $p_{r}$ and $p_{g}$. Noted that both $p_{r}$ and $p_{g}$ are low-dimensional manifolds in high-dimensional spaces, so to some extent, these two distribution are not not overlapped. In these cases, no matter how close the two distributions are, the JS divergence is $log2$ that cannot gradually update $G$ to generate better samples. Optimizing this objective function can easy lead to the problem of gradients vanishing during training process. So in practice, discriminator must synchronize well with generative network during training (in practice, we cannot train $G$ too much without updating $D$).
\subsection{EM distance driven GAN}

Minimizing the JS divergence in the context of optimizing distributions typically is not continuous, which leads to the vanishing gradients problem during the learning process. To solve this issue, Arjovsky et al. \cite{martin2017wasserstein} proposed Wasserstein GAN (WGAN) in 2017, which utilizes the EM distance $W(p_{r}; p_{g})$ as to measure the divergence between $p_{r}$ and $p_{g}$. EM distance is defined as the smallest average distance that the 'earth mover' has to move from $p_{g}$ to $p_{r}$. Theoretically, the EM distance $W(p_{r}; p_{g})$, under mild assumptions, is continuous and differentiable that might be a much more sensible cost function when optimized than at least JS divergence. Experiments in \cite{martin2017wasserstein,gulrajani2017improved} also illustrate WGAN can enhance the stability of learning and solve the problem of mode collapse to some extent.

The evaluation of Wasserstein distance between $p_{r}$ and $p_{g}$ in neural networks is:

\begin{footnotesize}
\begin{equation}
V(G,D)=\max_{D\epsilon 1-Lipschtiz} \{ \mathbb{E}_{x\sim p_{r}}[D(x)] -\mathbb{E}_{ \widetilde{x}\sim p_{g}} [D(\widetilde{x})] \}
\label{eq2}
\end{equation}
\end{footnotesize}

Where $D$ enforces the 1-Lipschtiz function that has a gradient with norm at most 1 everywhere.
An efficient method to enforce the Lipschitz constraint is gradient penalty \cite{gulrajani2017improved} that penalize the gradient norm on samples $\widehat{x} \sim p_{\widehat{x}}$ that presents the region between $p_{r}$ and $p_{g}$.
We can reformulate the objective as follows:
\begin{equation}
\begin{split}
V(G,D)\approx \max_{D} \{ \mathbb{E}_{x\sim p_{r}} [D(x)] - \mathbb{E}_{\widetilde{x}\sim p_{g}} [D(\widetilde{x})] \\
+\lambda \mathbb{E}_{\widehat{x} \sim p_{\widehat{x}}} [(||\nabla_{\widehat{x}}D(\widehat{x})||_{2} -1 )^{2} ]   \}
\end{split}
\label{eq3}\end{equation}

Given that the implementation of unit gradient norm constraints everywhere is tricky, $p_{\widehat{x}} $ only present the region between $p_{r}$ and $p_{g}$, because they influence the way $p_{g}$ moves to $p_{r}$, and the experimental results perform well.

\section{GAN-OVA}
\label{GAN-OVA}
\subsection{Model architecture}
In most GAN architectures, the output layer of the discriminator (or critic) is a one-dimensional scalar that can measure the fidelity of the generated image. Therefore, by optimizing the scalar of the generated image continuously, the generated image can be closer to the real image. In the conditional version of GANs, whether it is adding conditions to the input of the discriminator or judging conditions through an additional network, the discriminator still maintains this scalar output, and the primary function is to judge the confidence of a single case.
This scalar output structure of discriminator has demonstrated its effect on image synthesis, but for the conditional image synthesis, intuitively speaking, it may not fully release the capability of the discriminator.

In this paper, we explore a novel scheme for conditional image generation, namely GAN-OVA. The highlight of this structure is that we design a One-Vs-All classifier as a discriminator network to distinguish each condition.
Figure \ref{fig1} illustrates the framework's workflow. The generator network is the same as other conditional GANs, where noise $z$ and condition labels $c$ are both as generator's input to synthesize images $\widetilde{x} = G(z|c)$. The discriminator network is constructed with a One-Vs-All classifier with a vector output, which aims to distinguish each conditional sample from the generated images $\widetilde{x}$ and the real training images $x$. The real data $x$ has $N$ classes, $[c_{1},c_{2},...,c_{n}]$, according to the conditioned information $c$. The number of units in the One-Vs-All classifier's output layer depends on the number of types of condition $c$ and different divergences driven methods mentioned in Section \ref{Background}.

\subsection{Different methods driven GAN-OVA}
\begin{figure}[h]
  \centering
  \subfigure[]{
    %\label{fig:subfig:a} %% label for first subfigure
    \includegraphics[scale=0.55]{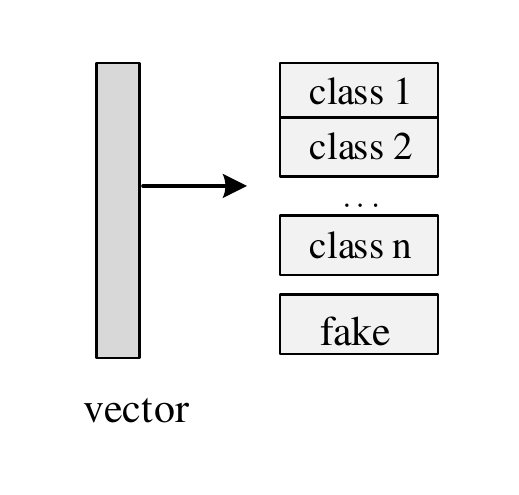}
  }
  \subfigure[]{
    %\label{fig:subfig:b} %% label for second subfigure
    \includegraphics[scale=0.55]{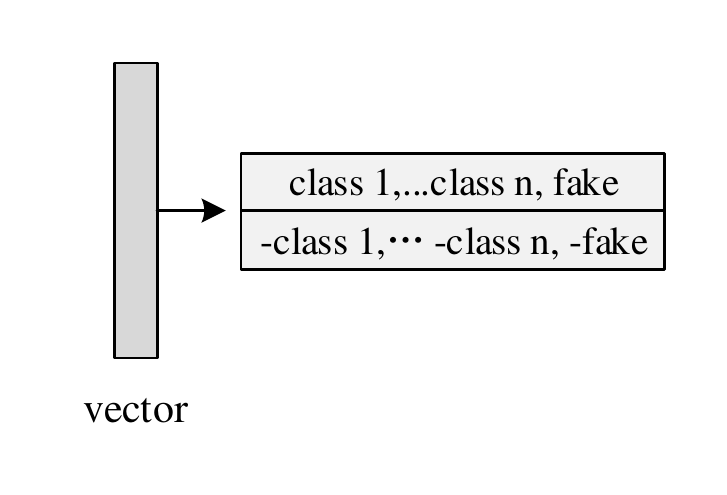}
  }
\caption{The target vector of different methods. (a): JS divergence-driven GAN-OVA; (b): EM distance-driven GAN-OVA.}
  %\label{fig:subfig} %% label for entire figure
\label{fig2}
\end{figure}

According to different model methods, the output layer and its target vector of One-Vs-All classifier can be designed as Figure \ref{fig2}.

In JS divergence-driven GAN, generator $G$ captures the data distribution, and discriminator $D$ serves as a binary classifier with the sigmoid cross entropy loss function to distinguish real data from generated samples. To extend the discriminator to  a multiclass situation, GAN-OVAs enlarge the units of discriminator's output layer and attach with the softmax cross entropy.

The One-Vs-All classifier gives the probability of $N+1$ classification labels, according to $[c_{1},c_{2},...,c_{n}, c_{fake}]$, where $c_{fake}$ denotes the label of synthesized images.

The objective functions would be given by the following equations:

\begin{footnotesize}
\begin{equation}
\begin{split}
\max_{D}J(D)=\mathbb{E}_{x\sim p_{r}}\sum_{j=1}^{n}y^{o}_{j}\ln(D(x))_{j} \\
+ \mathbb{E}_{\widetilde{x}\sim p_{g}}\sum_{j=1}^{n}y^{*}_{j}\ln D(G(c,z))_{j} \\
\max_{G}J(G)=\mathbb{E}_{\widetilde{x}\sim p_{g}}\sum_{j=1}^{n}y^{o}_{j}\ln D(G(c,z))_{j}
\end{split}
\label{eq4}
\end{equation}
\end{footnotesize}

Where $y^{o}$ denotes different classes of $p_{r}$, that is, $[c_{1},c_{2},...,c_{n}]$; $y^{*}$ denotes the class of the generated images, $c_{fake}$. Algorithm \ref{algorithm1} describes the training procedure shown in \ref{Algorithm}.

In EM distance driven GAN-OVA, the problem of finding the optimal EM distance between $P_{r}$ and $P_{g}$ can be seen as a maximization problem in solving equation \eqref{eq2}. Therefore, the target vector for real data is assigned to $[c_{1},c_{2},...,c_{n}]$ ($y^{o}$), and for the generated data is assigned to $[-c_{1},-c_{2},...-c_{n}]$ ($y^{*}$).
The objective can be described as follows:

\begin{footnotesize}
\begin{equation}
\begin{split}
V(G\!,\!D)\! \approx \max_{D}\! \{ \mathbb{E}_{x\sim p_{r}}\! \sum_{j=1}^{n}\! y^{o}_{j} \! D(x)_{j} \! - \! \mathbb{E}_{\widetilde{x}\sim p_{g}} \! \sum_{j=1}^{n}\! y^{*}_{j} \! D(\widetilde{x})_{j} \\
+\lambda \mathbb{E}_{\widehat{x} \sim p_{\widehat{x}}} [(||\nabla_{\widehat{x}}D(\widehat{x})||_{2} -1 )^{2} ]
\}
\end{split}
\label{eq5}\end{equation}
\end{footnotesize}

Here we also use the gradient penalty to enforce the 1-Lipschitz restriction, as in \cite{gulrajani2017improved}. $\widehat{x} \sim p_{\widehat{x}}$ presents the region between $p_{r}$ and $p_{g}$ used as gradient penalty object. Algorithm \ref{algorithm2} describes the training procedure shown in \ref{Algorithm}.

\section{Experiments}
\label{Experiments}
\vspace{-0.35cm}
In this section, the first goal is to verify the One-Vs-All classifier's effectiveness by assisting the conditioned generator to control the synthesis of samples on MNIST and CelebA-HQ datastes, and we discuss the model speed and sample quality by comparing with ACGAN. Then we test the GAN-OVA's stability by excluding the batch normalization (BN) for the generator, as in \cite{ioffe2015batch}. Finally, we explore the interpretable representation between the condition variables and the observations by manipulating different condition input.

\subsection {Datasets and Implementation Details}
We evaluate GAN-OVAs on two datasets, MNIST \cite{lecun1998gradient} and CelebA-HQ \cite{karras2017progressive}. The MNIST dataset contains 60,000 28 $\times$ 28 grayscale images in 10 hand-written classes of which 50,000 are as training set and 10,000 are as testing set. In the CelebA-HQ dataset, we split the dataset with two scenes, men and women, as different categories. Both man and woman images achieved over 10,000 images as training set and 1,000 images as testing set with samples resized to 64 $\times$ 64. The noise $z $ for all models comes from a uniform distribution of [-1,1] with dimension of 100.
The experiments were implemented in Tensorflow \cite{abadi2016tensorflow} with Python 3.6.

\subsection {The effectiveness of GAN-OVAs}

To demonstrate the effectiveness of conditional image generation, we train GAN-OVAs on MNIST dataset and CelebA-HQ face dataset. Our baseline comparison is standard ACGANs. For MNIST dataset, we use the simplest multilayer perceptron (MLP) as generator and discriminator structure, based on the JS divergence-driven method. Specifically, generator is an MLP with $4$ hidden layers of 256-512-1024-784 unites one after another, and discriminator is 4-layer MLP with 512 units at each layers. LeakyRelu \cite{dubey2019comparative} is applied both in training the generator network and One-Vs-All classifier, while dropout \cite{hinton2012improving} is applied in training the One-Vs-All classifier. The Adam optimizer \cite{kingma2014adam,wang2018deep} is used with parameters $\alpha = 0.0001,\beta_{1} = 0.,\beta_{2} = 0.9$.

For CelebA-HQ dataset, we keep the public DCGAN structure for GAN-OVA but exclude the batch normalization layers in the One-Vs-All classifier following \cite{mao2017least} based on the EM distance-driven method. Also use Adam optimizer. Detailed experimental setup is described at \ref{Architecture}.

\begin{figure}[htbp]
\centering

\subfigure[\tiny ACGAN 10 epochs]{
\begin{minipage}[t]{0.33\linewidth}
\centering
\includegraphics[width=1.02in]{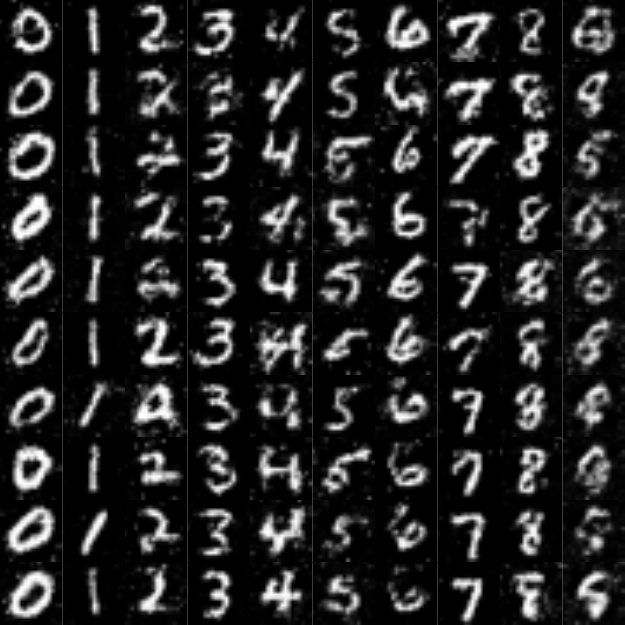}
%\caption{fig1}
\end{minipage}%
}%
\subfigure[\tiny ACGAN 30 epochs]{
\begin{minipage}[t]{0.33\linewidth}
\centering
\includegraphics[width=1.02in]{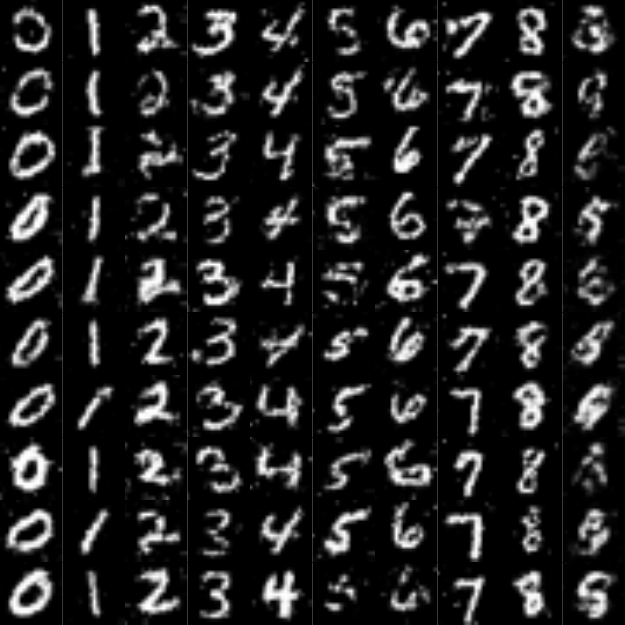}
%\caption{fig2}
\end{minipage}%
}%
\subfigure[\tiny ACGAN 50 epochs]{
\begin{minipage}[t]{0.33\linewidth}
\centering
\includegraphics[width=1.02in]{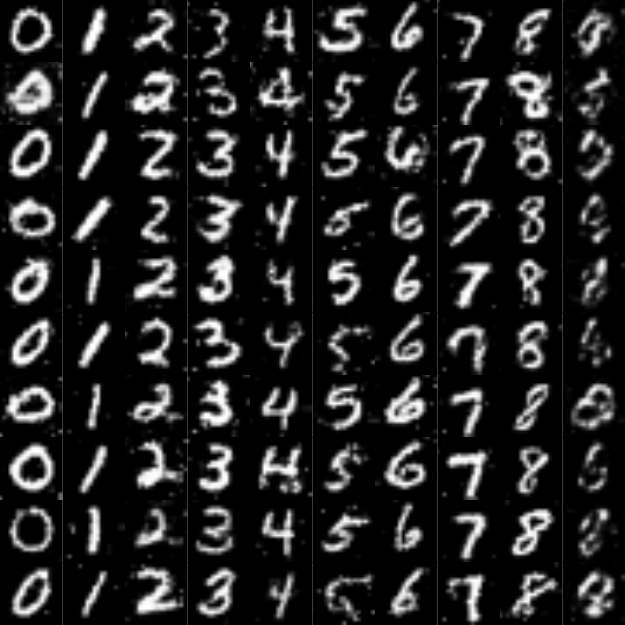}
%\caption{fig1}
\end{minipage}%
}%
\quad             %这个回车键很重要 \quad也可以
\subfigure[\tiny GAN-OVA 10 epochs]{
\begin{minipage}[t]{0.33\linewidth}
\centering
\includegraphics[width=1.02in]{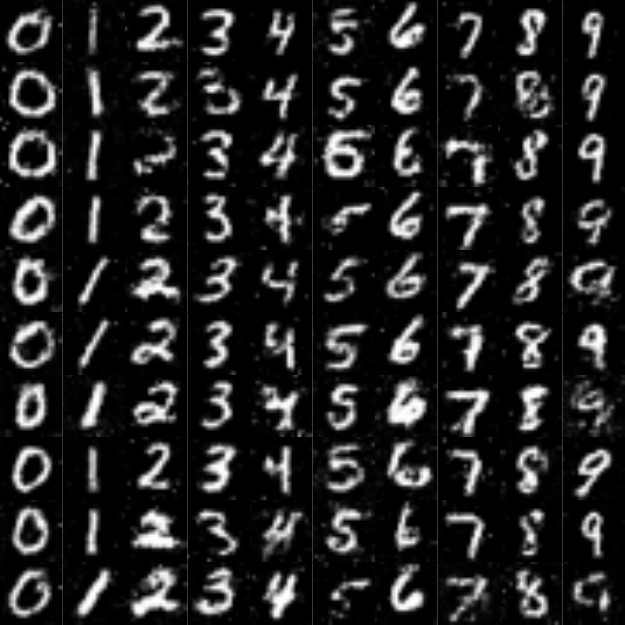}
%\caption{fig1}
\end{minipage}%
}%
\subfigure[\tiny GAN-OVA 30 epochs]{
\begin{minipage}[t]{0.33\linewidth}
\centering
\includegraphics[width=1.02in]{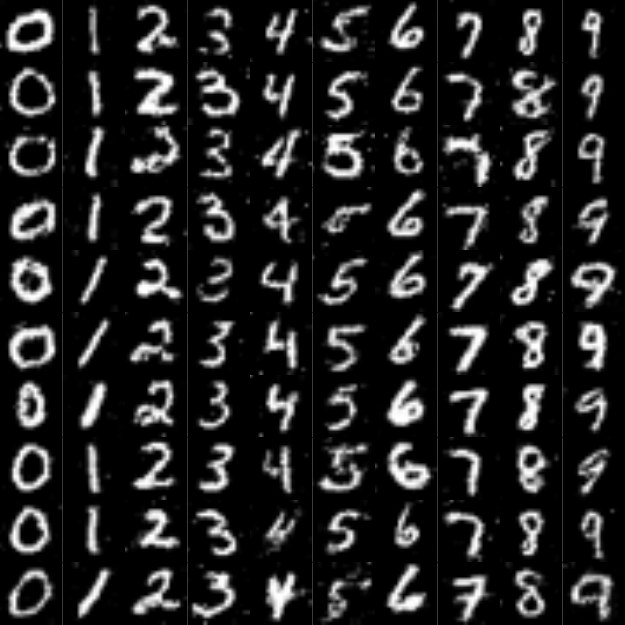}
%\caption{fig2}
\end{minipage}%
}%
\subfigure[\tiny GAN-OVA 50 epochs]{
\begin{minipage}[t]{0.33\linewidth}
\centering
\includegraphics[width=1.02in]{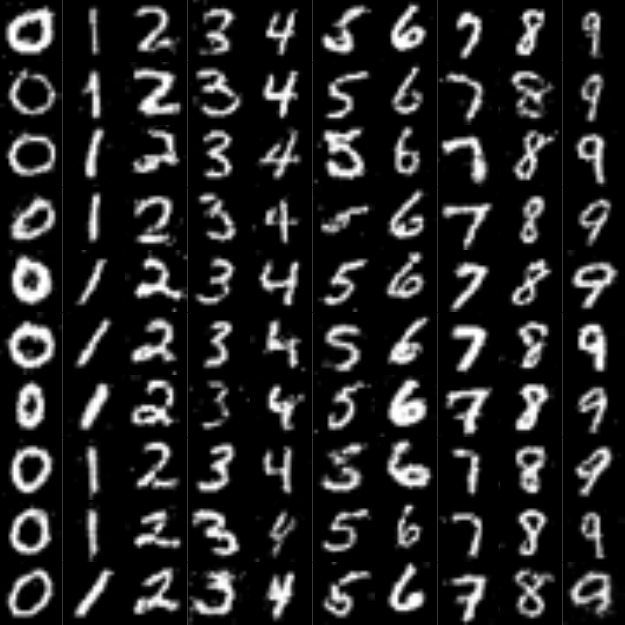}
%\caption{fig1}
\end{minipage}%
}%

\centering
\caption{Comparison of generated images in different epochs on the MNIST dataset. }
\label{fig3}
\end{figure}

Comparing the quality of the synthesized images of different models is challenging due to the lack of perceptible image similarity measures \cite{theis2015note} and one criteria does not mean good performance relative to other criteria.
Here, we use the most common visual fidelity as the indicator for generating image models on the MNIST dataset following \cite{li2018fast}.
Figure \ref{fig3} shows ACGAN and GAN-OVA images generated after 10, 30, and 50 epochs. By sampling different label categories, each column displays 0-9 digital images from left to right.
We observe that GAN-OVA can generate different type of digits corresponding to different conditions. Thus, we confirm that the One-Vs-All classifier successfully assists the conditioned generator to guide the synthesis of images. We also observe that GAN-OVA can achieve promising samples with 10 epochs while ACGAN achieve comparable results at 30 or more epochs. Therefore, we confirm that GAN-OVA effectively accelerates the generation process of different classes and improves the generation quality.

\begin{figure}[htbp]
\centering
\subfigure[\scriptsize man in ACGAN]{
\begin{minipage}[t]{0.45\linewidth}
\centering
\includegraphics[scale=0.43]{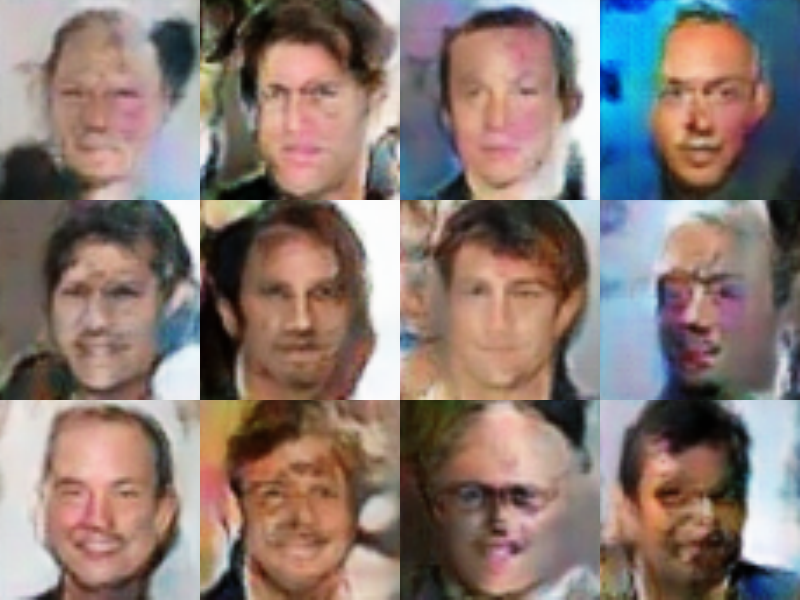}
%\caption{fig1}
\end{minipage}%
}%
\subfigure[\scriptsize woman in ACGAN]{
\begin{minipage}[t]{0.45\linewidth}
\centering
\includegraphics[scale=0.43]{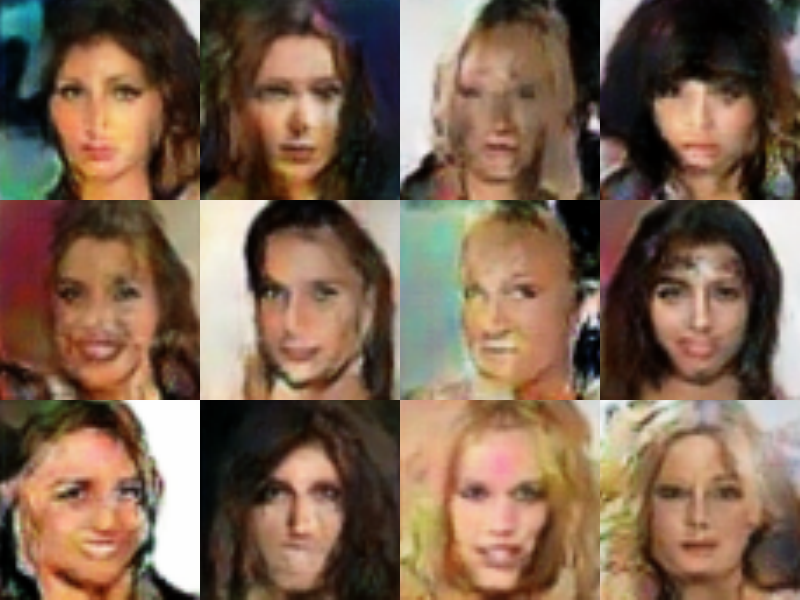}
%\caption{fig2}
\end{minipage}%
}%
\quad             %这个回车键很重要 \quad也可以
\subfigure[\scriptsize man in GAN-OVA]{
\begin{minipage}[t]{0.45\linewidth}
\centering
\includegraphics[scale=0.43]{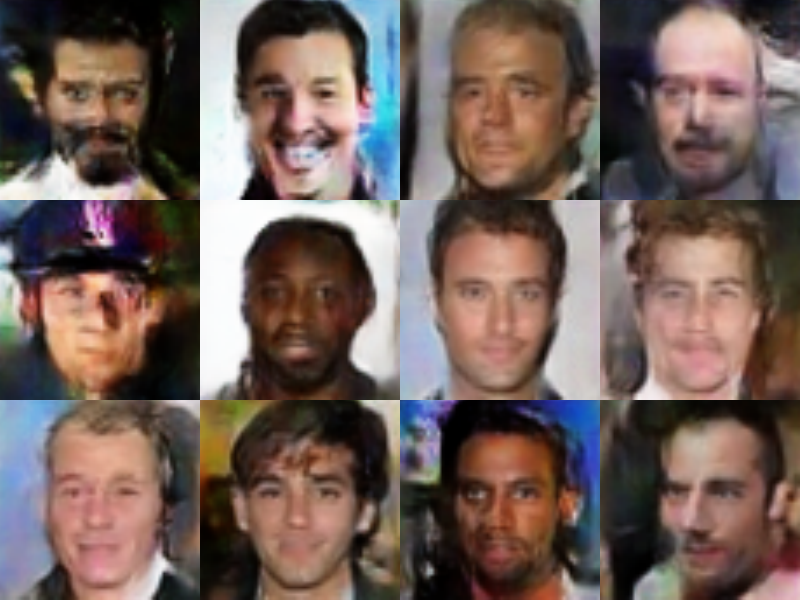}
%\caption{fig1}
\end{minipage}%
}%
\subfigure[\scriptsize woman in GAN-OVA]{
\begin{minipage}[t]{0.45\linewidth}
\centering
\includegraphics[scale=0.43]{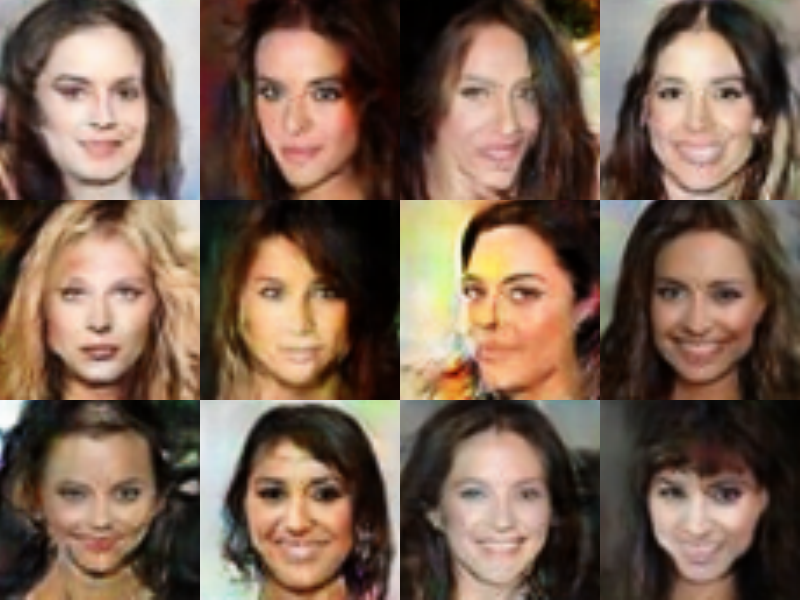}
%\caption{fig2}
\end{minipage}%
}%
\centering
\caption{Comparison of generated images on the CelebA-HQ dataset.}
\label{fig4}
\end{figure}

For CelebA-HQ, we train each model for 300K iterations and list samples in Figure \ref{fig4}.

We can see that images generated by GAN-OVAs are also clearer. Moreover, They contain more character details in hair and facial expressions.

\begin{figure}[htbp]
\centering
\includegraphics[scale=0.5]{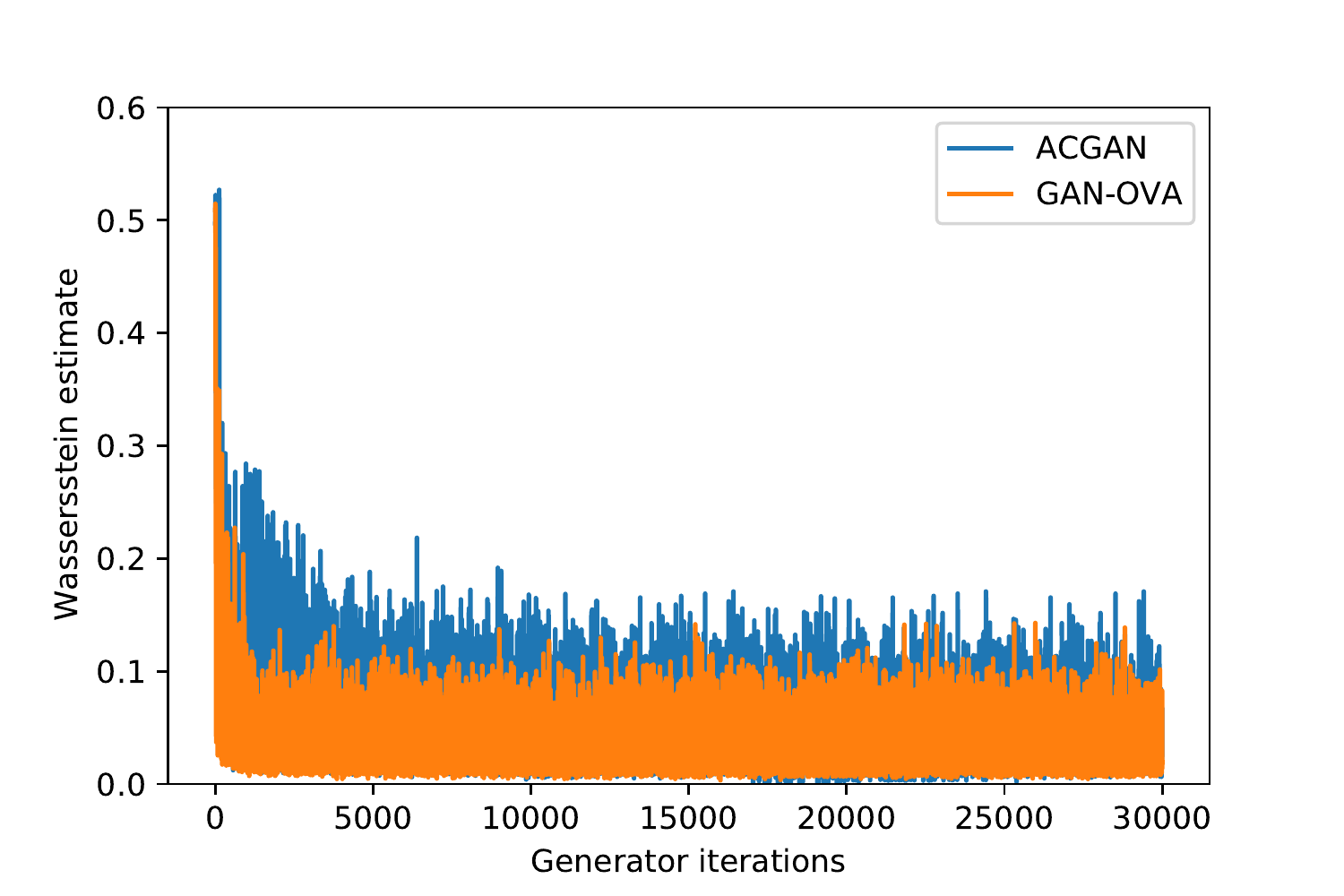}
\caption{The Wasserstein estimation at different stages of training.}
\label{fig5}
\end{figure}

To further compare the quality of the generated images, we measure the Wasserstein distance of GAN-OVA and ACGAN and present in Figure \ref{fig5}.

Wasserstein distance has proven to be a meaningful loss metric that correlates with the generator's convergence and sample quality \cite{martin2017wasserstein}. Compared with CGAN, GAN-OVA converges faster at the very beginning, and does a better estimation in a whole. Therefore, we confirm that GAN-OVA  improves the generation quality.

\subsection {Improved Stability}
\begin{figure}[htbp]
\centering

\subfigure[\scriptsize ACGAN]{
\begin{minipage}[t]{1\linewidth}
\centering
\includegraphics[scale=0.55]{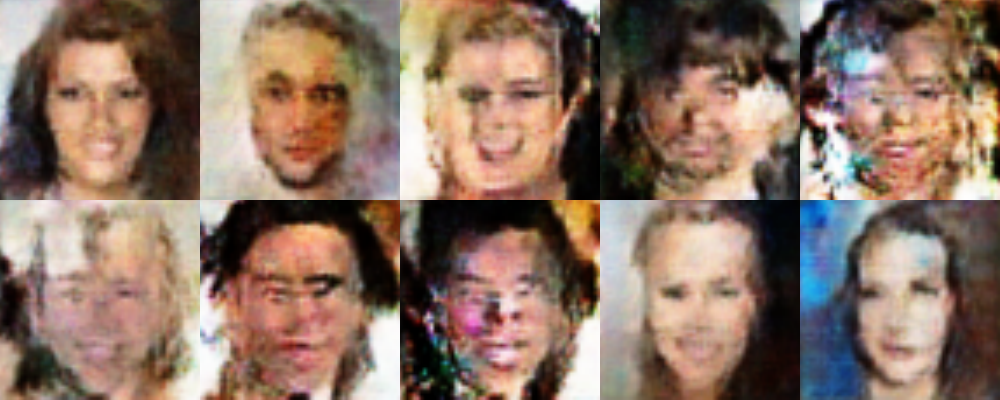}
%\caption{fig1}
\label{fig6a}
\end{minipage}%
}%

\quad
           %这个回车键很重要 \quad也可以
\subfigure[\scriptsize GAN-OVA]{
\begin{minipage}[t]{1\linewidth}
\centering
\includegraphics[scale=0.55]{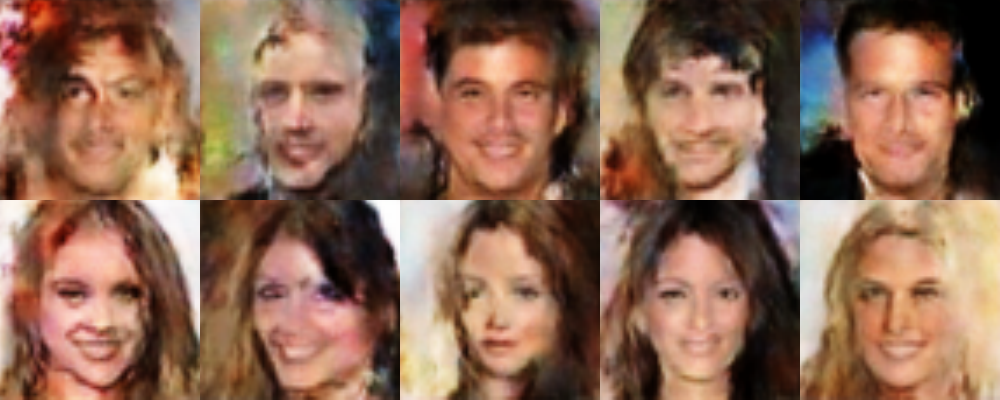}
%\caption{fig1}
\end{minipage}%
}%

\centering
\caption{ Comparison experiments by excluding batch normalization (BN). In each picture, the first row represents the generated images from man class and the second row represents the generated images from woman class. (a):GAN-OVA without BN in G. (b): ACGAN without BN in G.}
\label{fig6}
\end{figure}

The batch normalization layer is known as a trick to improve the stability of network training \cite{santurkar2018does,salimans2016improved}. Following \cite{mao2017least}, we only take off the BN for the generator and keep other settings to compare the stability between GAN-OVA and ACGAN on CelebA-HQ dataset. Figure \ref{fig6} shows the comparison results at 300k iterations. The first row of each picture represents images generated from the man category, and the second row represents images generated from the woman category.
we can see that both GAN-OVA and ACGAN suffer a little mode collapse, but GAN-OVA performs a more stable gradient than ACGAN. Specifically, in GAN-OVA, although some of the generated pictures lose some details, they are all recognizable. However, excluding BN in ACGAN overwhelmingly degrades the quality of the generated images, and make them blurred or even lose their category identification. Take the first row of Figure \ref{fig6a} as an example, the first image is definitely in a wrong label, and following four images are difficult to discern their class type. Therefore, we can confirm that GAN-OVA improves the model stability.

\subsection {Interpretable representations}
In this section, we discussed the interpretable representation \cite{chen2016infogan} between the condition variables and the observations. Both datasets of MNIST and CelebA-HQ are trained on DCGAN architecture with EM distance-driven method and keep other default settings.

\begin{figure*}[htbp]
\centering
\includegraphics[scale=0.4]{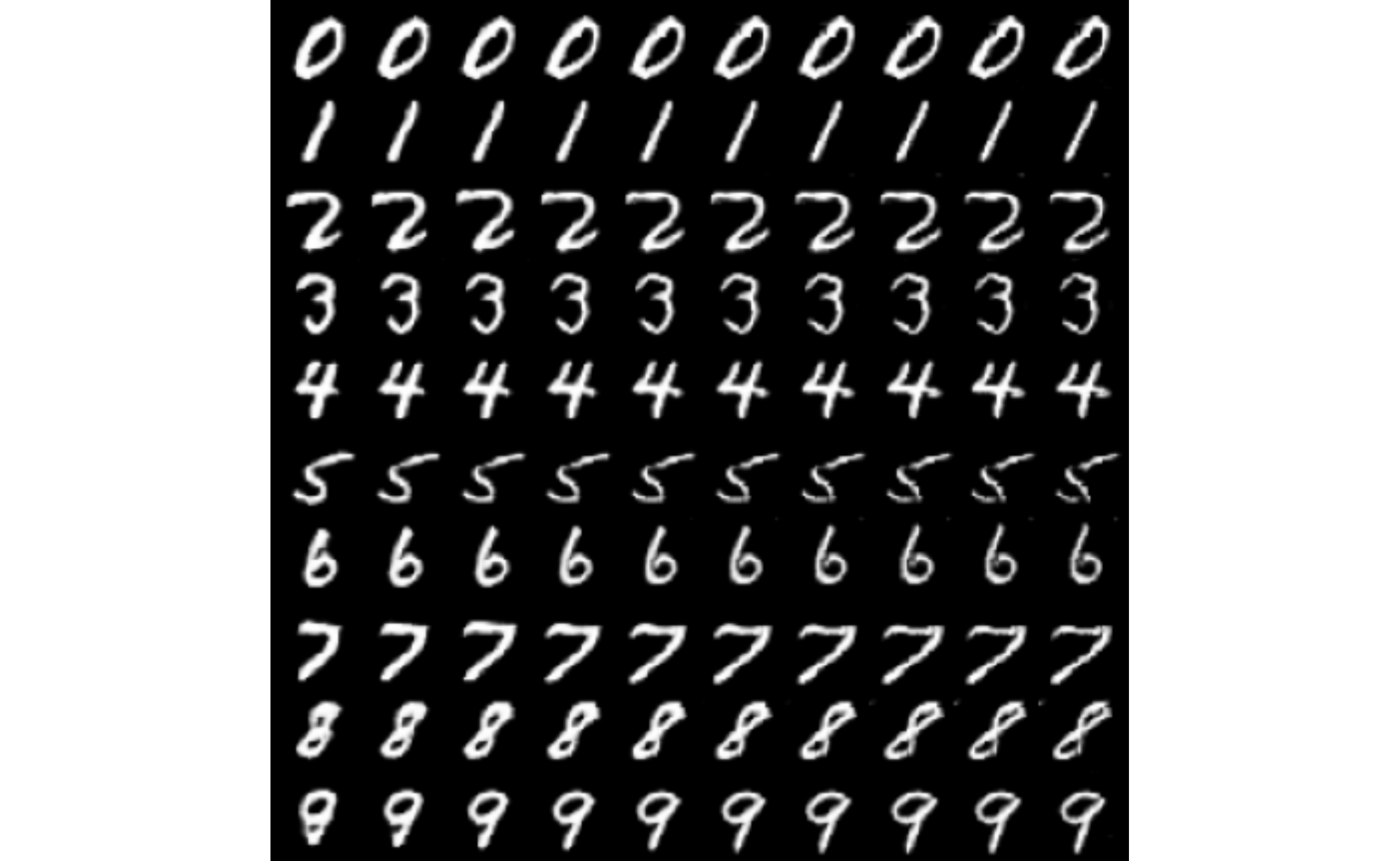}
\caption{Manipulating condition codes on MNIST dataset: In all manipulations, each one condition code is varied from left to right, while the other condition codes and noise are fixed. The different rows correspond to different type of digits.}
\label{fig7}
\end{figure*}

Figure \ref{fig7} shows the interpretable results on MNIST dataset. The conditional inputs in generator are used by the images class labels encoded as one-hot codes, in which each condition code is serves as a digit type.

We only varied one condition code at a time to assess if varying such code results in style variation of the corresponding digit type. From left to right, one condition code value is varied from 0.5 to 1.85. It shows that the digits smoothly change from wide to narrow, which means the value of each condition code controls the thickness of the digit's stroke.

\begin{figure*}[htbp]
\centering
\includegraphics[scale=0.4]{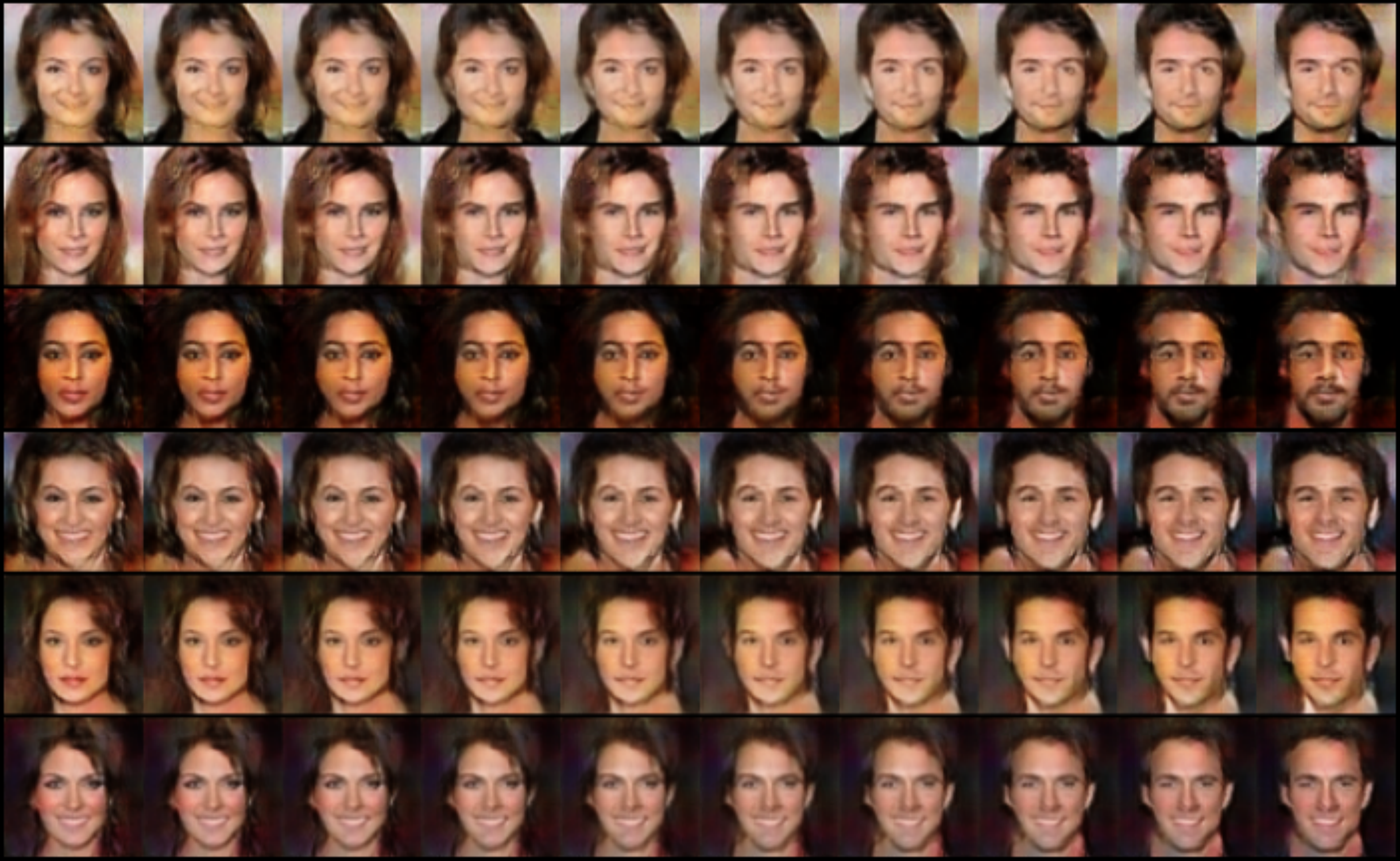}
\caption{ The effect of the learned conditions on the outputs on CelebA-HQ dataset as their values vary from [0,1] to [1,0].}
\label{fig8}
\end{figure*}

The interpretable results on CelebA-HQ dataset are shown in Figure \ref{fig8}. Here we change the conditional input from [0,1] to [1,0] with fixed noise. It shows that as the condition values varying, the output images are smoothly change from woman to man, which means each condition code has successfully affect the outputs type.
Therefore, we can confirm that each condition code has learned the data representation.
\section{Conclusion and discussion}
\label{Conclusion and discussion}
This work proposes the GAN-OVA architecture for conditional image generation. Instead of the real/fake discriminator used in vanilla GANs, we propose expanding the discriminator as a One-Vs-All classifier (GAN-OVA) that can distinguish each input data to their class label.  Our model can be applied to different types of divergences used to evaluate the divergence between generated data and the real data, such as Jensen-Shannon divergence and Earth-Mover distance. Experimental results show that GAN-OVAs ont only make progress toward stable training, but also effectively accelerate the generation process of different classes and improves generation quality.

Based on the current findings, we hope to explore GAN-OVA to achieve more reliable modeling performance in other domains, such as video \cite{kumar2019videoflow}, audio \cite{vasquez2019melnet} and language \cite{yin2013icdar}.
\section*{Acknowledgment}
This work was supported by the National Natural Science Foundation of China (grant number F060609).
\bibliography{mybibfile}
\clearpage
\appendix
\renewcommand\thefigure{\Alph{section}\arabic{figure}}
\section{Architecture}
\label{Architecture}

\begin{table}[htb]
\centering
\caption{The GAN-OVA architecture for MNIST in interpretable representation testing.}
\resizebox{\textwidth}{17mm}{
\begin{tabular}{rccc|cccc}
\hline
\multicolumn{4}{c}{Generator}& \multicolumn{4}{c}{Discriminator} \\
\hline
Layer & Filter/Stride & Resample &Output Size& Layer & Filter/Stride &Resample&Output Size\\
\specialrule{0em}{1pt}{1pt}
Linear,Relu&-&- & $128\times7\times7$                   &Conv,LeakyRelu & 3$\times$3/2& -& 16$\times$14$\times$14\\
Conv,Relu  & 4$\times$4/1 & Up &128$\times$14$\times$14 &Conv,LeakyRelu & 3$\times$3/2& -& 32$\times$7$\times$7\\
Conv,Relu  & 4$\times$4/1 & Up &64$\times$28$\times$28  &Conv,LeakyRelu & 3$\times$3/1& -  &128$\times$4$\times$4\\
Conv,Tanh  & 4$\times$4/1 & - &1$\times$28$\times$28 &Linear& -& -&10\\
LeakyRelu slope&\multicolumn{6}{l}{0.2}\\
Optimizer&\multicolumn{6}{l}{ Adam ( $ \alpha=0.0001, \beta_{1}=0, \beta_{2}=0.9$) }\\
\hline
\end{tabular}}
\setlength{\abovecaptionskip}{0pt}
\label{table A.1}
\end{table}

\begin{table}[htb]
\centering
\caption{The GAN-OVA architecture for CelebA-HQ.}
\resizebox{\textwidth}{26mm}{
\begin{tabular}{rccc|cccc}
\hline
\multicolumn{4}{c}{Generator}& \multicolumn{4}{c}{Discriminator} \\
\hline
Layer & Filter/Stride & BN? &Output Size& Layer & Filter/Stride &BN?&Output Size\\
\specialrule{0em}{1pt}{1pt}
Input1&4$\times$4/1& $\surd$ & $256\times4\times4$    &Conv, LeakyReLu & 5$\times$5/2& $\times$& 128$\times$32$\times$32\\
Input2& 4$\times$4/1 & $\surd$ &256$\times$4$\times$4 &Conv, LeakyReLu & 5$\times$5/2& $\times$& 256$\times$16$\times$16\\
Deconv, relu & 4$\times$4/2 & $\surd$ &256$\times$8$\times$8 &Conv, LeakyReLu & 5$\times$5/2& $\times$ &256$\times$8$\times$8\\
DeConv, relu & 4$\times$4/2 & $\surd$ &128$\times$16$\times$16 &Conv, LeakyReLu& 3$\times$3/2& $\times$&512$\times$4$\times$4\\
DeConv, relu & 4$\times$4/2 & $\surd$ &128$\times$32$\times$32 &Conv, LeakyReLu& 5$\times$5/4& $\times$&512$\times$1$\times$1\\
DeConv, tanh & 4$\times$4/2 & $\surd$ &3$\times$64$\times$64 &Linear& -& $\times$ &2\\
LeakyReLu slope&\multicolumn{7}{l}{ 0.2 }\\
Optimizer&\multicolumn{7}{l}{ Adam( $ \alpha=0.0001, \beta_{1}=0, \beta_{2}=0.9$) }\\
BN ?& \multicolumn{7}{l}{Indicates whether the layer is followed by a batch normalization layer}\\
BN momentum& \multicolumn{7}{l}{ 0.8 }\\
Weight,bias init& \multicolumn{7}{l}{Isotropic gaussian ($\mu$ = 0, $\sigma$ = 0.02), Constant(0)}\\
\hline
\end{tabular}}
\label{table A.2}
\end{table}
\section{Algorithm}
\label{Algorithm}
\begin{algorithm*}[ht]
\caption{JS divergence-driven GAN-OVA. Here we used $k = 1$. The batch size $m = 100$. $y^{o}$ denotes the labels for real data and $y^{*}$ denotes the labels for generated data.}
  \begin{algorithmic}[1]
    \For{each training iteration}
      \For {k steps}
      \State Sample $m$ samples $\{(c^{(i)},x^{(i)})\}_{i=1}^{m}$ from the real data distribution $p_{r}(x)$.
      \State Sample $m$ noise samples $\{z^{(i)}\}_{i=1}^{m}$ from a noise prior $p_{z}(z)$.
      \State Obtaining generated data $\{\widetilde{x}^{(i)}\}_{i=1}^{m},\ \widetilde{x}^{(i)}= G(c^{(i)},z^{(i)}) $.
      \State Update discriminator parameters $\theta_{d}$ to maximize.
      \State $ \widetilde{V} = \frac{1}{m}\sum_{i=1}^{m}\sum_{j=1}^{n} {y^{o}}_{j}^{(i)} \ln(D(x^{(i)}))_{j} +\frac{1}{m}\sum_{i=1}^{m} \sum_{j=1}^{n} {y^{*}}_{j}^{(i)}\ln (1 - D(\widetilde{x}^{(i)}))_{j} $
      \State $\theta_{d} \leftarrow \theta_{d} + \eta \nabla\widetilde{V}(\theta_{d}) $
      \EndFor
      \State Sample $m$ noise samples $\{z^{(i)}\}_{i=1}^{m}$ from a noise prior $p_{z}(z)$.
      \State Sample $m$ conditions $\{c^{(i)}\}_{i=1}^{m}$ from the data distribution $p_{r} (x)$.
      \State Update generator parameters  $\theta_{g} $to maximize.
      \State $ \widetilde{V} = \frac{1}{m} \sum_{i=1}^{m}\sum_{j=1}^{n} {y^{o}}_{j}^{(i)} \ln(D(c^{(i)},z^{(i)}))_{j}$
      \State $ \theta_{g} \leftarrow \theta_{g} - \eta\nabla\widetilde{V}(\theta_{g})$
    \EndFor
  \end{algorithmic}
  \label{algorithm1}
\end{algorithm*}
\begin{algorithm*}[hp]
\caption{EM distance-driven GAN-OVA. Here we used $k = 5$. The batch size $m = 100$. $y^{o}$ denotes the labels for real data and $y^{*}$ denotes the labels for generated data.}
 \begin{algorithmic}[1]
    \For{each training iteration}
      \For{k steps}
        \State Sample $m$ examples $\{(c^{(i)},x^{(i)})\}_{i=1}^{m}$ from the real data distribution $p_{r} (x)$.
        \State Sample $m$ noise samples $\{z^{(i)}\}_{i=1}^{m} $ from a noise prior $p_{z}(z)$.
        \State $\widetilde{x}^{i} \leftarrow G(c^{(i)},z^{(i)})$
        \State $\widehat{x}^{i} \leftarrow \varepsilon x^{(i)} + (1-\varepsilon) \widetilde{x}^{i} $
        \State Update discriminator parameters $\theta_{d}$ to maximize.

        \State $ \widetilde{V} = \frac{1}{m} \sum_{i=1}^{m}\sum_{j=1}^{n} {y^{o}}^{(i)}_{j} D(x)_{j} - \frac{1}{m}\sum_{i=1}^{m}\sum_{j=1}^{n} {y^{*}}_{j}^{(i)}D(\widetilde{x}^{i})_{j} +\lambda [\| \nabla_{\hat{x}} D_{\theta_{d}(\hat{x})}\|_{2}-1]^{2} $
        \State $\theta_{d} \leftarrow \theta_{d} + \eta \nabla\widetilde{V}(\theta_{d}) $

      \EndFor
      \State Sample $m$ noise samples $\{z^{(i)}\}_{i=1}^{m} $  from a noise prior $p_{z}(z)$.
      \State Sample $m$ conditions $\{c^{(i)}\}_{i=1}^{m}$ from the data distribution $p_{r} (x)$.
      \State Update generator parameters  $\theta_{g} $to maximize.
      \State $ \widetilde{V} = \frac{1}{m}\sum_{i=1}^{m}\sum_{j=1}^{n}{y^{o}}_{j}^{(i)}D(G(c^{i},z^{i}))_{j} $
      \State $ \theta_{g} \leftarrow \theta_{g} - \eta\nabla\widetilde{V}(\theta_{g})$

    \EndFor
    \label{code:recentEnd}
  \end{algorithmic}
  \label{algorithm2}
\end{algorithm*}
\end{document}